\documentclass[runningheads]{llncs}
\usepackage{graphicx}
\usepackage{multirow}
\usepackage{subcaption}
\usepackage{xcolor}
\usepackage{comment}
\begin{document}
\title{Explainable Landscape Analysis in Automated Algorithm Performance Prediction}
%
\titlerunning{Explainable Landscape Analysis}
%
\author{Risto Trajanov\inst{1} \and
Stefan Dimeski\inst{1}\and
Martin Popovski\inst{1}\and
Peter Koro\v{s}ec\inst{2}\orcidID{0000-0003-4492-4603} \and
Tome Eftimov\inst{2}\orcidID{0000-0001-7330-1902}}
\authorrunning{R. Trajanov et al.}
%
\institute{Faculty of Computer Science and Engineering, Ss. Cyril and Methodius, University -
Skopje, North Macedonia\\
\email{risto.trajanov@gmail.com,\\\{stefan.dimeski.1,martin.popovski\}@students.finki.ukim.mk}\\
 \and
Computer Systems Department, Jo\v{z}ef Stefan Institute, Ljubljana, Slovenia
\email{\{peter.korosec,tome.eftimov\}@ijs.si}}
\maketitle              
\begin{abstract}

Predicting the performance of an optimization algorithm on a new problem instance is crucial in order to select the most appropriate algorithm for solving that problem instance. For this purpose, recent studies learn a supervised machine learning (ML) model using a set of problem landscape features linked to the performance achieved by the optimization algorithm. However, these models are black-box with the only goal of achieving good predictive performance, without providing explanations which landscape features contribute the most to the prediction of the performance achieved by the optimization algorithm. In this study, we investigate the expressiveness of problem landscape features utilized by different supervised ML models in automated algorithm performance prediction. The experimental results point out 
that the selection of the supervised ML method is crucial, since different supervised ML regression models utilize the problem landscape features differently and there is no common pattern with regard to which landscape features are the most informative.

\end{abstract}
\section{Introduction}
Automated algorithm performance prediction plays a crucial part in the automated algorithm selection and configuration tasks~\cite{BlotMJH19,HutterKV19,JankovicD20,KerschkeT19,LiefoogheDVDAT20}. The most common practices are to train a supervised machine learning (ML) model using a set of problem landscape features. The ML model links the characteristics of the problem instance landscape to the performance achieved by an optimization algorithm that is run on that instance. However, such ML models are still black-box with a limited explanations of how each landscape feature of the problem instance influences the prediction of the end performance result achieved by an optimization algorithm. 

To describe the characteristics of a problem instance, the Exploratory Landscape Analysis(ELA) ~\cite{mersmann2011exploratory} is used, where for each problem instance a set of landscape features are calculated, known as ELA features. These features are coming from different groups (e.g., statistical, information theory, etc.) and require a selection of a sampling technique and a sample size that will be used for their calculation. They can be split into two groups, cheap and expensive, based on the computational time required to calculate them.  

The idea behind the automated algorithm performance prediction is to link the problem instances landscape data to the performance data achieved by an optimization algorithm. For this purpose, the algorithm is run on a set of benchmark problem instances (i.e., in most cases on already defined benchmark suite such as COCO~\cite{hansen2020coco}) to collect the performance data. Next, the ELA features are calculated in order to describe the characteristics of the problem instances. Finally, in order to predict the performance of the algorithm on a new problem instance, a supervised ML method is trained using the landscape data as input data and the performance data as a target data. In recent studies, this is done by learning a single ML model using a set of ELA features that works well across all problem instances~\cite{jankovic2021towards,jankovic2021impact,kerschke2019automated}. All these studies used classical feature selection methods to select the ELA features that should improve the performance of the ML model. However, the key element missing here is the explainability of the ML performance, or providing explanations which ELA features contribute to the end performance prediction. This kind of analysis is more than needed to understand which ELA features are the most informative ones and can be used to make a good algorithm performance prediction. These empirical insights will also provide new directions for theoretical research. Even more, it has been shown that using different supervised ML methods with the same landscape and performance data provide different results~\cite{jankovic2021impact}. Therefore, the selection of the ML algorithm depends on the optimization algorithm whose performance data is used as a target.

In this paper, we present a ML pipeline that can be utilized to understand the expressiveness of the ELA features in automated algorithm performance prediction. The main contribution of the paper is that the ELA feature importance changes when different supervised ML methods are utilized, so their expressiveness on the automated algorithm performance prediction is questionable. It depends from the problem instance being solved, the optimization algorithm run on that problem instance, and the supervised ML methods used to learn a predictive model. 


\section{Related work}
There are two different types of studies where the ELA features are utilized:
\begin{itemize}
    \item Studies performed only in the landscape space.
    \item Studies performed to link the landscape data to the performance data.
\end{itemize}

In the first type of studies, the ELA features are used to describe the problem instances and then these representations are used to perform complementary analysis between different sets of problem instances~\cite{lang2021exploratory,skvorc2020}. With this kind of analysis, similar problem instances can be detected. In addition, sensitivity analysis of the ELA features are preformed concerning different sampling techniques and sample sizes, where it has been shown that the ELA features are really sensitive to the sampling techniques and sample sizes that are used for their calculation~\cite{renau2020exploratory,vskvorc2021effect}. Different ELA features portfolios have been also investigated in order to see if the information they convey is enough to classify each instance to the problem to which belongs~\cite{eftimov2020linear,renau2021towards}. The common thing of all above-mentioned analyses is that all are done only in the landscape space, and the relations with the performance space have not been explored.

The second type of studies involve automated algorithm performance prediction as a regression task. Here, it has been shown that different supervised ML regression models provide different results when they are utilized for the same learning task~\cite{jankovic2021impact}. So depending on which optimization algorithm is used in the  prediction, different supervised ML model should be selected for learning the model. In addition, it has been shown that personalizing the regression models to the problem instance that is being solved can decrease the predictive error~\cite{eftimov2021personalizing}. Furthermore, a recent study provides global (across all benchmark problem instances) and local (for a single problem instance) explanations of which ELA features are most important when a supervised ML algorithm is used to predict an optimization algorithm performance~\cite{trajanov2021explainable}. However, these explanations have not been analyzed when different supervised ML methods are used in the predictive task. This analysis is extremely important to investigate if there is some pattern showing which ELA features are the most informative when automated algorithm performance prediction is investigated no matter which supervised ML model is used.

\section{Automated algorithm performance prediction}
Previous studies have already shown that models trained to predict the target precision reached by an algorithm or its logarithmic value perform differently~\cite{jankovic2021towards,jankovic2021impact}. There are problem instances for which the model trained to predict the target precision works well, however there are also problem instances for which the model trained to predict the logarithmic value of the target precision works better. To decide which model should be used, several previous studies have analyzed different empirical thresholds for switching between the original and the logarithmic model. To analyze the importance of the ELA features in automated algorithm performance prediction, we investigated different supervised ML regression methods (decision trees (DT), random forest (RF), and deep neural network (DNN)) in Single Target Regression(STR) and Multi Target Regression(MTR) learning scenario. Two STR models will be investigated (i.e., one per the target precision and one per its logarithmic value) together with one MTR model that predicts both values (i.e., the target precision and its logarithmic value) simultaneously. This will be explored using the three supervised ML regression methods.
 
 \section{Experimental setup}
Next, the experimental setup is explained in more detail providing information about the landscape and performance data. In addition, the utilized supervised ML models together with their hyper-parameters are presented in more detail.
\subsection{Data}
\subsubsection{Landscape data} The COCO benchmark platform ~\cite{hansen2020coco}, comprised of 24 single-objective continuous optimization problems, is selected to represent the problem space.
For the experiments presented here, the problem dimension is set to $D=5$ and the first 50 instances per problem are included. Because of this experimental design, there are 1,200 problem instances, 50 per each problem. The R package ``flacco"~\cite{kerschke2017package} has been used for the calculation of the landscape characteristics (i.e. ELA features) of the problem instances. Out of all the calculated ELA features, 99 have been selected. The calculated ELA features are from the following groups: \emph{cm\_angle}, \emph{cm\_grad}, \emph{disp}, \emph{ela\_conv}, \emph{ela\_curv}, \emph{ela\_distr}, \emph{ela\_level}, \emph{ela\_local}, \emph{ela\_meta}, \emph{ic}, and \emph{nbc}. The calculation of the selected ELA features has been done using the improved latin hypercube sampling method ~\cite{xu2017improved} utilizing $50D$ sample points. This process has been repeated 10 times for each problem instance and each ELA feature is actually the median value of its 10 repetitions. The reason for choosing the median over the mean is that the median value is more statistically robust than the mean value. Regarding the computation cost, the selected ELA features are the cheap ones and they do not have any missing values.

\subsubsection{Performance data} One randomly selected configuration for modular CMA-ES has been investigated as performance data. The selected CMA-ES configuration has the following hyper-parameters: Active update = FALSE, Elitism  = TRUE, Orthogonal Sampling = TRUE, Sequential selection = FALSE, Threshold Convergence = TRUE, Step Size Adaptation = tpa, Mirrored Sampling = mirrored, Quasi-Gaussian Sampling = halton, Recombination Weights = default, Restart Strategy = BIPOP.~\cite{nobel2020} contains additional information about the hyper-parameters of the modular CMA-ES. Only one randomly selected configuration has been presented as a proof of concept about the analysis, but in our GitHub repository~\cite{data}, there are results for another 14 CMA-ES configurations, which makes this analysis a personalized task. Each CMA-ES configuration has been run 10 times on each problem instance in a fixed budget scenario, where the budget has been set to 50,000 function evaluations. In our analysis, the focus of the performance prediction is on the best reached target precision, thus as a final result, the median across all 10 runs of the best reached precision has been selected.
The target variables in our case are the target precision and its logarithmic transformation with a slight modification: before base 10 logarithm is performed, one is added to each original target precision reached. The purpose of this modification is for getting a better interpretation of the performance measure that is used for evaluating the regression models. 

\subsection{Regression models and their hyper-parameters}
Here, the three regression models (i.e., Decision Tree(DT), Random Forest(RF), and Deep Neural Network(DNN)) together with their hyper-parameters are explained in more detail. The DT and RF have been selected because recent studies~\cite{jankovic2020landscape,jankovic2021impact} showed that they provide one of the most promising results. The selected hyper-parameters for both DT and RF are summarized in Table~\ref{tab:hyperparameters_regression}. For both models, ``\textit{mae}" has been selected to measure the quality of the split. The maximum depth of the DT in Multi Target Regression(MTR) scenario has been set to 10, while in Single Target Regression(STR) is set to 9. The $max\_depth$ of the DT in STR scenario is set to 9 so that the maximum possible sum of the number of leaf nodes of the two DT used in STR is equal to the maximum possible number of leaf nodes of the DT in the MTR scenario, since the trees are binary. In case of RF, the number of trees in the forest in the MTR scenario has been set to 20, while in STR is set to 10. The maximum depth of the tree is set to 7 in both scenarios. Since in-depth hyper-parameter analysis has not been performed for this study, the number of trees in the MTR RF scenario is actually the sum of trees that appear in the STR RF models. 

\begin{table}[tb]
    \centering
     \caption{Hyper-parameter values for tree-based regression models.}
     \label{tab:hyperparameters_regression}
    \begin{tabular}{l@{\hskip 0.1in}l}
    \hline
    \multicolumn{1}{c}{Algorithm} & \multicolumn{1}{c}{Hyper-parameters}\\
    \hline
        \multirow{2}{*}{Decision Tree} & $crit = ``mae"$,  \\
          & $max\_depth = 10\ (MTR)\ and\ max\_depth = 9\ (STR) $ \\
         \hline
        \multirow{3}{*}{Random Forest} &  $crit = ``mae"$ \\
     &  $max\_depth = 7\ (MTR\ and\ STR)$ \\
         & $n\_estimators=20\ (MTR)\ and\ n\_estimators=10\ (STR)  $ \\
        \hline
    \end{tabular}
\end{table}

\begin{figure}[ht]
\begin{subfigure}[b]{0.5\linewidth}
    \centering
    \includegraphics[width=1.\textwidth]{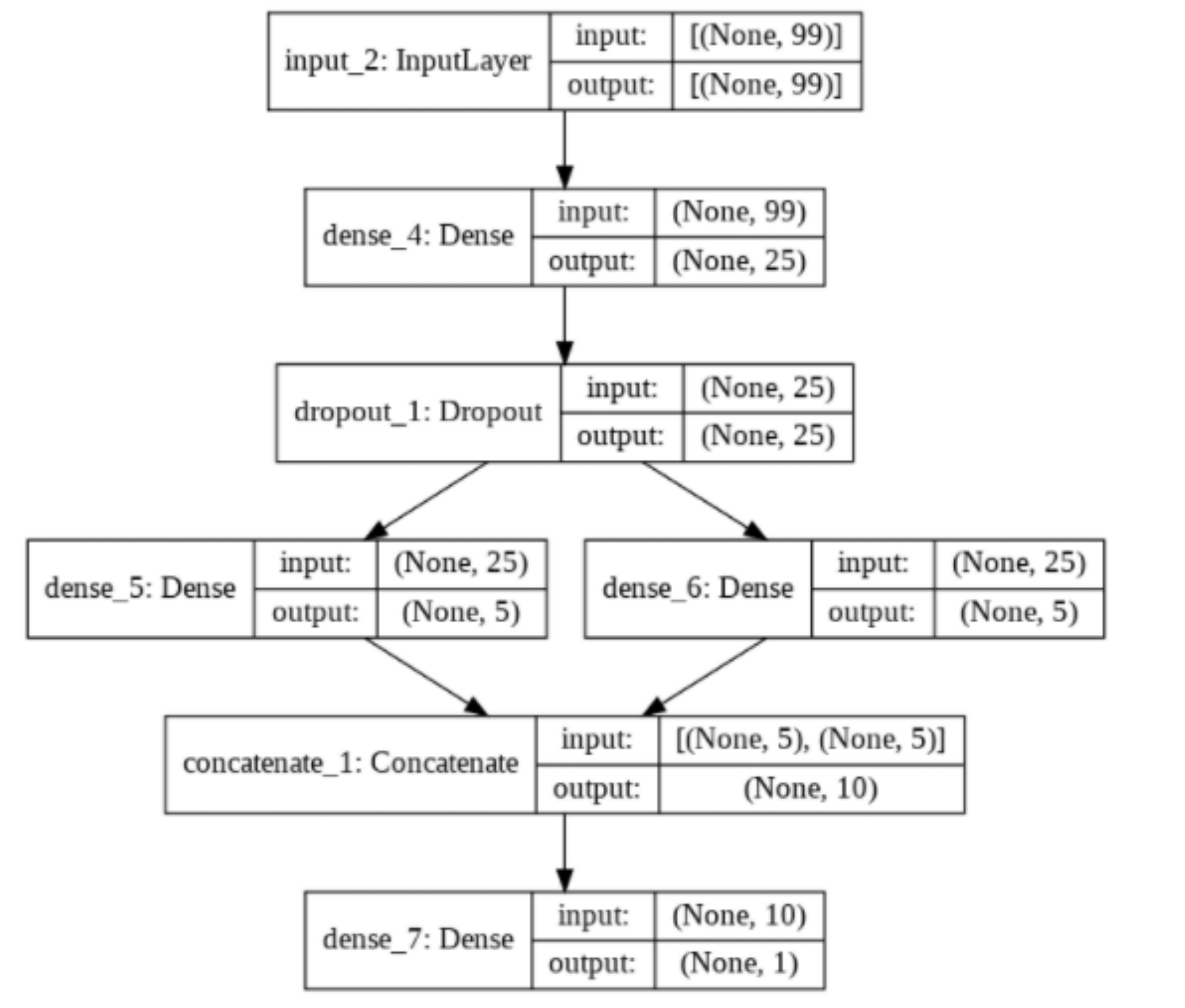}
    \caption{STR learning scenario.} 
    \label{fig:dnn_str} 
    \vspace{4ex}
  \end{subfigure}
\begin{subfigure}[b]{0.5\linewidth}
    \centering
    \includegraphics[width=1.\textwidth]{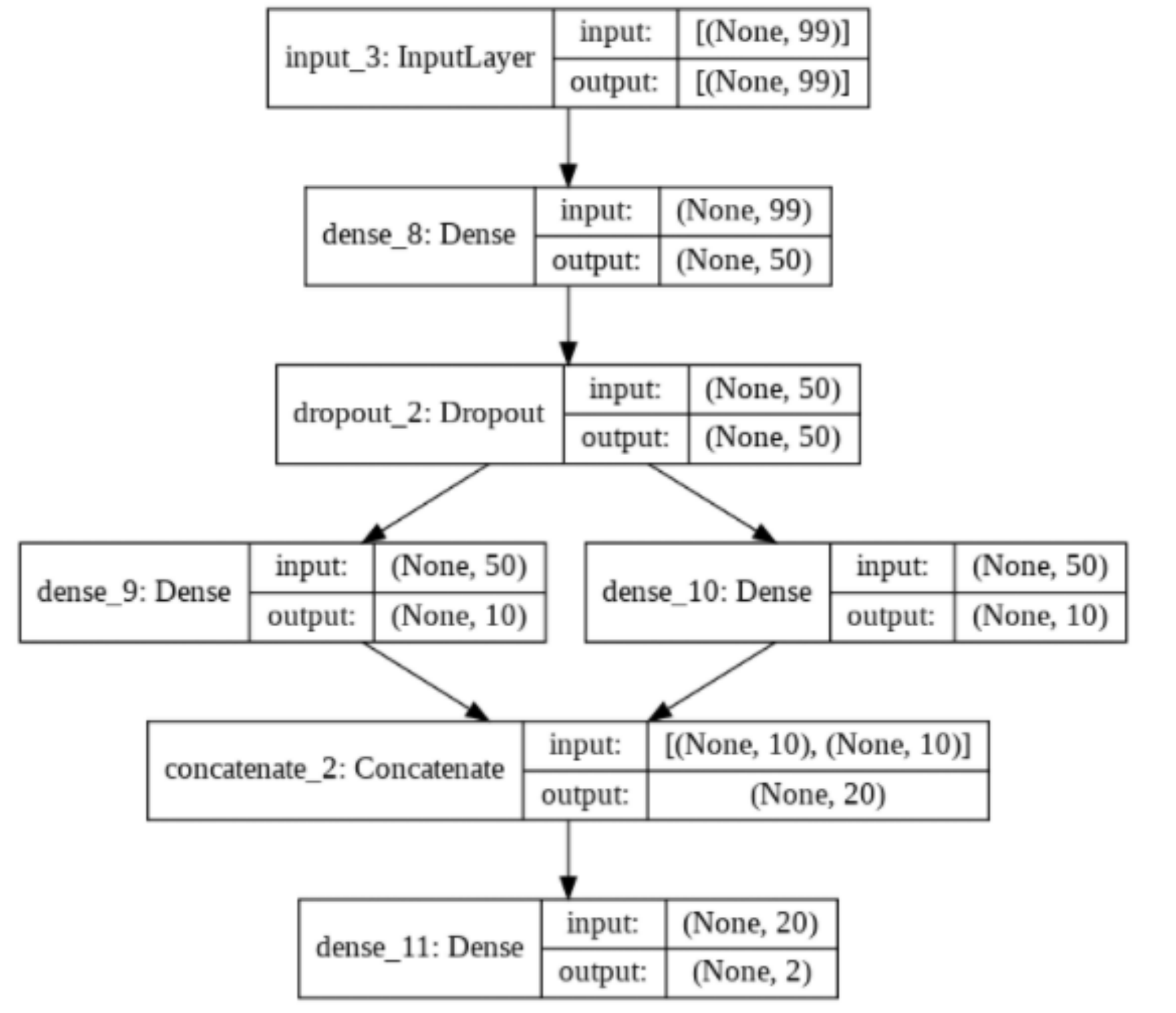}
    \caption{MTR learning scenario.} 
    \label{fig:dnn_mtr}
    \vspace{4ex}
  \end{subfigure}
  \caption{DNN architectures.}
\end{figure}

The tested DNN in STR scenario is presented in Figure~\ref{fig:dnn_str}, while the DNN for MTR is presented in Figure~\ref{fig:dnn_mtr}. We have decided for similar architectures to test the prediction results, where finding the best neural architecture design has not been the focus of this study. The focus of this study is to investigate the ELA features importance that are utilized by different supervised ML methods used for the same learning task. The designs of the DNNs used in STR and MTR scenarios are very similar, they have the same layer structure (i.e., seven layers) and use the same activation functions in the corresponding layers. In terms of activation functions, the hidden dense layers all use the ReLU activation function, while the output layer uses the linear activation function.

There are several main differences between the DNNs' design for the two scenarios:
\begin{itemize}
  \item The dense, concatenate, and dropout layers in the STR scenario all have exactly half of the number of inputs and outputs as the corresponding layers in the MTR scenario. The only exception to this is the first layer after the input layer, because it has the same number of inputs in both scenarios, since the same features are used.
  \item The output layer of each of the two DNNs in the STR scenario has one output neuron, while the output layer of the DNN in the MTR scenario has two output neurons.
  \item The number of parameters is 2,771 for STR and 6,062 for MTR (roughly twice of the number of parameters used in STR).
\end{itemize}

Both DNNs were trained for 100 epochs each, using the Adam optimizer with $learning\_rate$ set to 0.001 and $batch\_size$ set to 10.

\subsection{Evaluation} The Single Target Regression(STR) and Multi Target Regression(MTR) models are trained and evaluated using 50-fold cross validation. The data set consisting of 1,200 problem instances has been split into 50 folds such that the first fold contains the first instances of all 24 problems, the second contains the second instances of all 24 problems and so on till the 50th fold which contains the 50th instances of all 24 problems. The learning process was repeated fifty times, each time using one of the folds for testing and the others for training the regression model. This evaluation follows the idea of leave-one problem-instance out, since leaving all instances of one problem out does not provide promising results and does not transfer the knowledge learned by the model. This comes from the fact that the existing benchmark problem suites are not representative enough to generalize a performance of a ML model, which is not the focus of this study. The selected benchmark suite has been used to provide explanations about the expressiveness of the ELA features.

\section{Results and discussion}
\label{sec:results}

Tables~\ref{tab:DT_RF_models} and \ref{tab:DNN_models} present the mean absolute error (MAE) across all 50 folds for each of the 24 benchmark problem separately obtained by DT, RF, and DNN models in STR and MTR learning scenarios.  From the results, there is no general conclusion if the MTR models are better than the STR models, so there are benchmark problems for which STR models provide better mean absolute error, and vice versa. In general, the STR models obtained by the DT models provide better MAE  across all benchmark problems (i.e., 25.088 and 0.180) than the MTR models (i.e, 27.950 and 0.208) for the target precision and its logarithmic transformation, respectively. In case of the RF models, the MTR model provides better MAE (i.e., 21.381) across all benchmark problems than the STR model learned for the original target precision reached (i.e., 21.559). The opposite is true when the prediction is done for the logarithmic transformation of the original target. The same holds true for the DNN models in the STR and MTR scenarios.

\begin{table}[ht]
\centering
\footnotesize
\caption{Mean absolute error across the 50 folds for each COCO benchmark problem obtained by DT and RF models in STR and MTR scenario.}
\label{tab:DT_RF_models}
\begin{tabular}{lr@{\hskip 0.1in}r@{\hskip 0.1in}rr@{\hskip 0.1in}r@{\hskip 0.1in}r@{\hskip 0.2in}r@{\hskip 0.1in}r@{\hskip 0.1in}rr@{\hskip 0.1in}r}
   \hline
\multicolumn{1}{c}{\multirow{3}{*}{f}} &\multicolumn{5}{c}{\textrm{DT}} && \multicolumn{5}{c}{\textrm{RF}}\\
  \cline{2-6}\cline{8-12}
& \multicolumn{2}{c}{\textrm{target}} && \multicolumn{2}{c}{\textrm{log\_target}} && \multicolumn{2}{c}{\textrm{target}} && \multicolumn{2}{c}{\textrm{log\_target}}\\
  \cline{2-3}\cline{5-6} \cline{8-9} \cline{11-12}
 & \multicolumn{1}{c}{STR} & \multicolumn{1}{c}{MTR} && \multicolumn{1}{c}{STR} & \multicolumn{1}{c}{MTR}  && \multicolumn{1}{c}{STR} & \multicolumn{1}{c}{MTR} && \multicolumn{1}{c}{STR} & \multicolumn{1}{c}{MTR} \\
  \hline

1    & 0.074             & \textbf{0.057}   && 0.013            & \textbf{0.012}   && 1.095             & \textbf{0.461}     && \textbf{0.012}   & 0.013           \\ 
2    & \textbf{132.867}  & 139.609          && \textbf{0.348}   & 0.405            && 106.448           & \textbf{97.127}    && 0.346            & \textbf{0.335}  \\ 
3    & \textbf{4.105}    & 4.198            && \textbf{0.106}   & 0.146            && 2.775             & \textbf{1.942}     && \textbf{0.105}   & 0.117           \\ 
4    & \textbf{1.605}    & 1.824            && \textbf{0.073}   & 0.082            && \textbf{3.236}    & 3.499              && \textbf{0.068}   & 0.101           \\ 
5    & 0.000             & \textbf{0.000}   && 0.014            & \textbf{0.000}   && 0.577             & \textbf{0.561}     && \textbf{0.007}   & 0.019           \\ 
6    & \textbf{6.897}    & 7.393            && \textbf{0.034}   & 0.143            && \textbf{2.449}    & 3.699              && \textbf{0.084}   & 0.125           \\ 
7    & \textbf{105.273}  & 141.233          && \textbf{1.041}   & 1.218            && \textbf{113.143}  & 118.557            && 0.987            & \textbf{0.964}  \\ 
8    & \textbf{0.719}    & 0.807            && \textbf{0.074}   & 0.121            && \textbf{0.688}    & 0.839              && \textbf{0.092}   & 0.114           \\ 
9    & 6.740             & \textbf{6.671}   && \textbf{0.069}   & 0.101            && \textbf{1.629}    & 2.972              && \textbf{0.058}   & 0.077           \\ 
10   & \textbf{137.634}  & 138.199          && 0.463            & \textbf{0.441}   && \textbf{98.529}   & 100.714            && 0.358            & \textbf{0.340}  \\ 
11   & \textbf{39.892}   & 51.325           && \textbf{0.301}   & 0.483            && \textbf{39.745}   & 41.054             && \textbf{0.324}   & 0.410           \\ 
12   & \textbf{139.217}  & 152.809          && \textbf{0.343}   & 0.356            && 109.999           & \textbf{106.186}   && 0.265            & \textbf{0.256}  \\ 
13   & \textbf{4.340}    & 4.638            && 0.193            & \textbf{0.192}   && \textbf{3.912}    & 3.976              && \textbf{0.162}   & 0.164           \\      
14   & 1.416             & \textbf{1.269}   && \textbf{0.050}   & 0.089            && \textbf{0.747}    & 0.968              && \textbf{0.061}   & 0.146           \\  
15   & 1.436             & \textbf{1.086}   && 0.090            & \textbf{0.082}   && 5.527             & \textbf{2.260}     && \textbf{0.084}   & 0.111           \\ 
16   & \textbf{1.254}    & 1.374            && 0.133            & \textbf{0.126}   && 2.571             & \textbf{1.916}     && \textbf{0.130}   & 0.216           \\ 
17   & \textbf{3.446}    & 3.545            && \textbf{0.174}   & 0.213            && \textbf{4.418}    & 4.960              && \textbf{0.126}   & 0.265           \\ 
18   & 11.109            & \textbf{10.196}  && 0.280            & \textbf{0.209}   && \textbf{13.109}   & 13.503             && \textbf{0.237}   & 0.351           \\ 
19   & 0.195             & \textbf{0.055}   && 0.022            & \textbf{0.019}   && \textbf{1.170}    & 1.627              && \textbf{0.050}   & 0.207           \\ 
20   & 0.289             & \textbf{0.231}   && \textbf{0.036}   & 0.037            && 1.091             & \textbf{0.388}     && \textbf{0.039}   & 0.077           \\    
21   & \textbf{0.593}    & 0.709            && \textbf{0.152}   & 0.167            && \textbf{0.652}    & 1.684              && \textbf{0.135}   & 0.143           \\ 
22   & \textbf{0.658}    & 0.776            && \textbf{0.166}   & 0.178            && 1.423             & \textbf{0.911}     && 0.153            & \textbf{0.150}  \\ 
23   & 0.549             & \textbf{0.459}   && 0.047            & \textbf{0.036}   && 0.418             & \textbf{0.411}     && \textbf{0.033}   & 0.042           \\    
24   & \textbf{1.816}    & 2.349            && \textbf{0.102}   & 0.148            && \textbf{2.065}    & 2.929              && \textbf{0.105}   & 0.122           \\                      
   \hline
   Mean & \textbf{25.089} &	27.951 &&	\textbf{0.180}&	0.209 && 21.559 &	\textbf{21.381} &&	\textbf{0.168} &	0.203\\
   \hline
\end{tabular}
\end{table}

\begin{table}[!hbt]
\centering
\footnotesize
\caption{Mean absolute error across the 50 folds for each COCO benchmark problem obtained by DNN models in STR and MTR scenarios.}
\label{tab:DNN_models}
\begin{tabular}{lr@{\hskip 0.1in}r@{\hskip 0.1in}r@{\hskip 0.1in}r}
  \hline
\multicolumn{1}{c}{\multirow{2}{*}{f}} & \multicolumn{2}{c}{\textrm{target}} & \multicolumn{2}{c}{\textrm{log\_target}} \\
  \cline{2-5}
 & \multicolumn{1}{c}{STR} & \multicolumn{1}{c}{MTR} & \multicolumn{1}{c}{STR} & \multicolumn{1}{c}{MTR} \\
  \hline
1  & 0.559        & \textbf{0.292} & \textbf{0.048} & 0.078          \\
2  & \textbf{92.219}  & 93.970           & \textbf{0.327}   & 0.332            \\
3  & 2.228          & \textbf{1.756}   & \textbf{0.126}  & 0.158           \\
4  & \textbf{2.399}  & 3.055              & \textbf{0.087}  & 0.140           \\
5  & 0.175          & \textbf{0.123} & \textbf{0.026} & 0.051          \\
6  & 1.199& \textbf{0.435} & 0.094          & \textbf{0.092} \\
7  & 90.217           & \textbf{89.238}  & 0.911            & \textbf{0.904}  \\
8  & 1.126           & \textbf{0.665} & 0.116           & \textbf{0.100}  \\
9  & 0.644          & \textbf{0.538} & \textbf{0.072}  & 0.092           \\
10 & \textbf{88.878}  & 92.989           & \textbf{0.307}   & 0.393           \\
11 & 5.737           & \textbf{2.176}  & 0.231        & \textbf{0.181}  \\
12 & \textbf{105.375}   & 106.930           & \textbf{0.417}  & 0.417            \\
13 & 6.862           & \textbf{5.478}  & \textbf{0.228}  & 0.335           \\
14 & \textbf{1.938}   & 2.193           & \textbf{0.169}  & 0.451           \\
15 & 2.468           & \textbf{1.594}  & 0.135         & \textbf{0.130}   \\
16 & 2.361           & \textbf{2.009}   & \textbf{0.156}  & 0.202           \\
17 & \textbf{3.100}  & 3.492           & 0.176         & \textbf{0.175}  \\
18 & 14.605           & \textbf{14.400}  & \textbf{0.293}  & 0.364           \\
19 & 0.614          & \textbf{0.309} & \textbf{0.037} & 0.077          \\
20 & 0.953          & \textbf{0.640} & \textbf{0.067}  & 0.119           \\
21 & 0.609          & \textbf{0.578} & \textbf{0.135}  & 0.143           \\
22 & \textbf{0.632} & 0.736          & \textbf{0.148}  & 0.163           \\
23 & \textbf{0.493} & 0.495          & \textbf{0.060}  & 0.085           \\
24 & 2.339           & \textbf{2.289}  & \textbf{0.138}  & 0.172 \\
\hline
   Mean      & 17.822                   & \textbf{17.766}            & \textbf{0.186}                & 0.223 \\ 
   \hline
\end{tabular}
\end{table}

Evaluating across the models, the DNN MTR model provides the best MAE (i.e., 17.765) across all benchmark problems for the original target precision, while the RF STR model provides the best MAE for predicting the logarithmic transformation of the target precision reached (i.e., 0.167). The obtained results show us that there is no practical difference in investigating the performance prediction (i.e., original and its logarithmic transformation) in STR and MTR scenario. The only benefit could be the time required to train the models, instead of training two STR models, one MTR model can provide very similar results.

Comparing the models on a single-problem level trained for prediction the target precision, it is obvious that all models in STR (DT, RF, and DNN) obtain large errors on the 2nd, 7th, 10th, 12th, and 18th problem. However for the 11th problem, both DT and RF provide large errors, while the DNN provides an error which is much more smaller than the the other two models.

No matter which ML model has been utilized in the STR or MTR learning scenario, the results are similar. The focus of this study is actually to go into more explanations by providing which ELA features are used by different models to provide their predictions. For this purpose, the SHAP method is used to provide explanations on global and local level~\cite{NIPS2017_7062}. The SHAP explanation method computes Shapley values from coalitional game theory with the goal of explaining the prediction of a single instance. It is a method to explain individual predictions. It does this explanation by computing the contribution of each feature to the prediction. The feature values of a data instance play the role of players in a coalition. Shapley values tell us how to fairly distribute the ``payout" (i.e., the prediction) among the features. Two levels of explanations are provided by the SHAP method:
\begin{itemize}
    \item Global explanations - the SHAP values for each problem instance are used to study the impact to the target variable (the impact across all problem instances involved in the learning process). This is done by combining the SHAP values for each problem instance.
    \item Local explanations – allow us to study the similarities/differences in the importance of the features for different problem instances (we can see how the feature importance changes across different problem instances). This is done through the use of the SHAP values for each problem instance.  Each problem instance gets its own set of SHAP values.
\end{itemize}

In our experiments, the explanations are provided for the models trained on the first fold and a clustering analysis of the models is presented across all folds based on the Shapley values of the ELA features.

\subsubsection{Global explanations}
\begin{figure*}[ht] 
  \begin{subfigure}[b]{0.5\linewidth}
    \centering
    \includegraphics[width=0.75\linewidth]{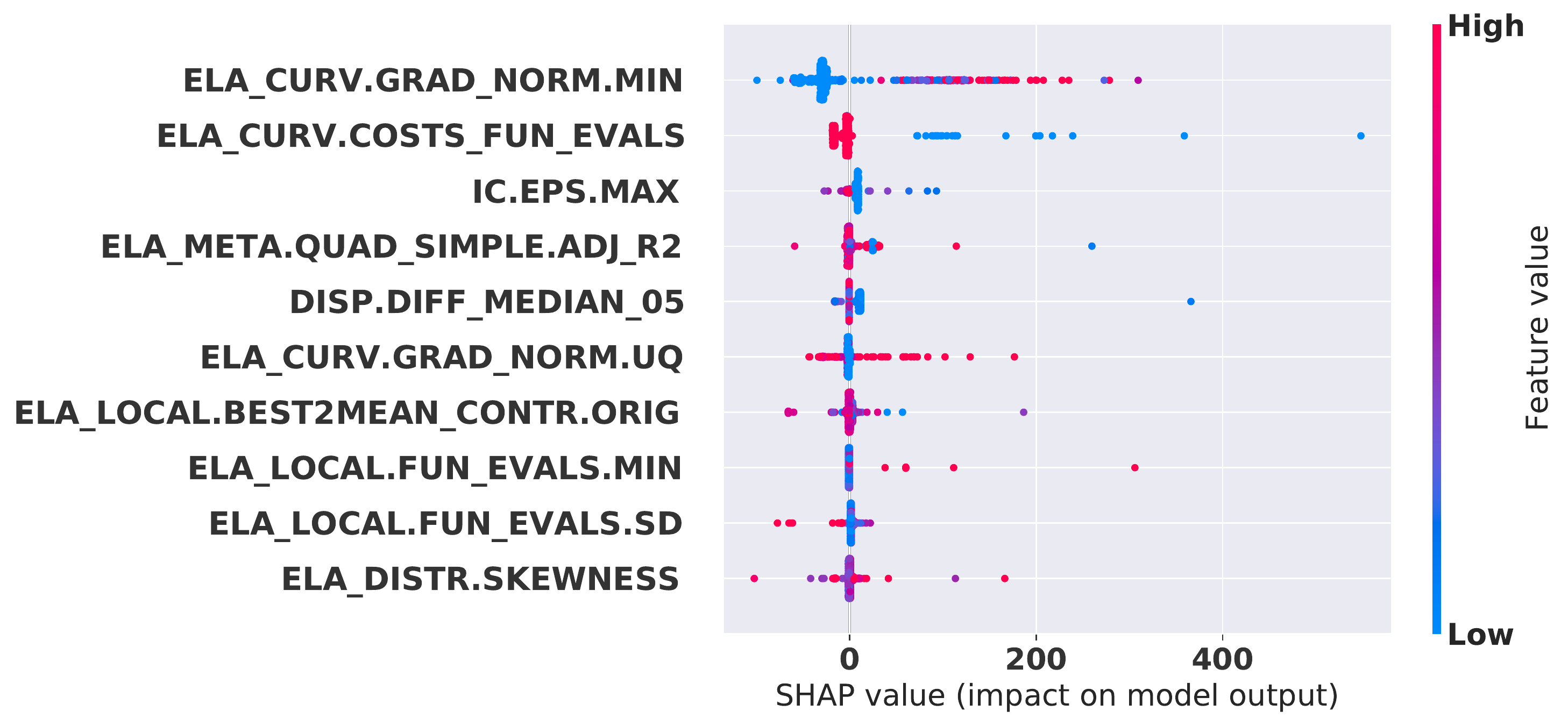} 
    \caption{Target precision STR DT.} 
    \label{fig7:a} 
    \vspace{4ex}
  \end{subfigure}
  \begin{subfigure}[b]{0.5\linewidth}
    \centering
    \includegraphics[width=0.75\linewidth]{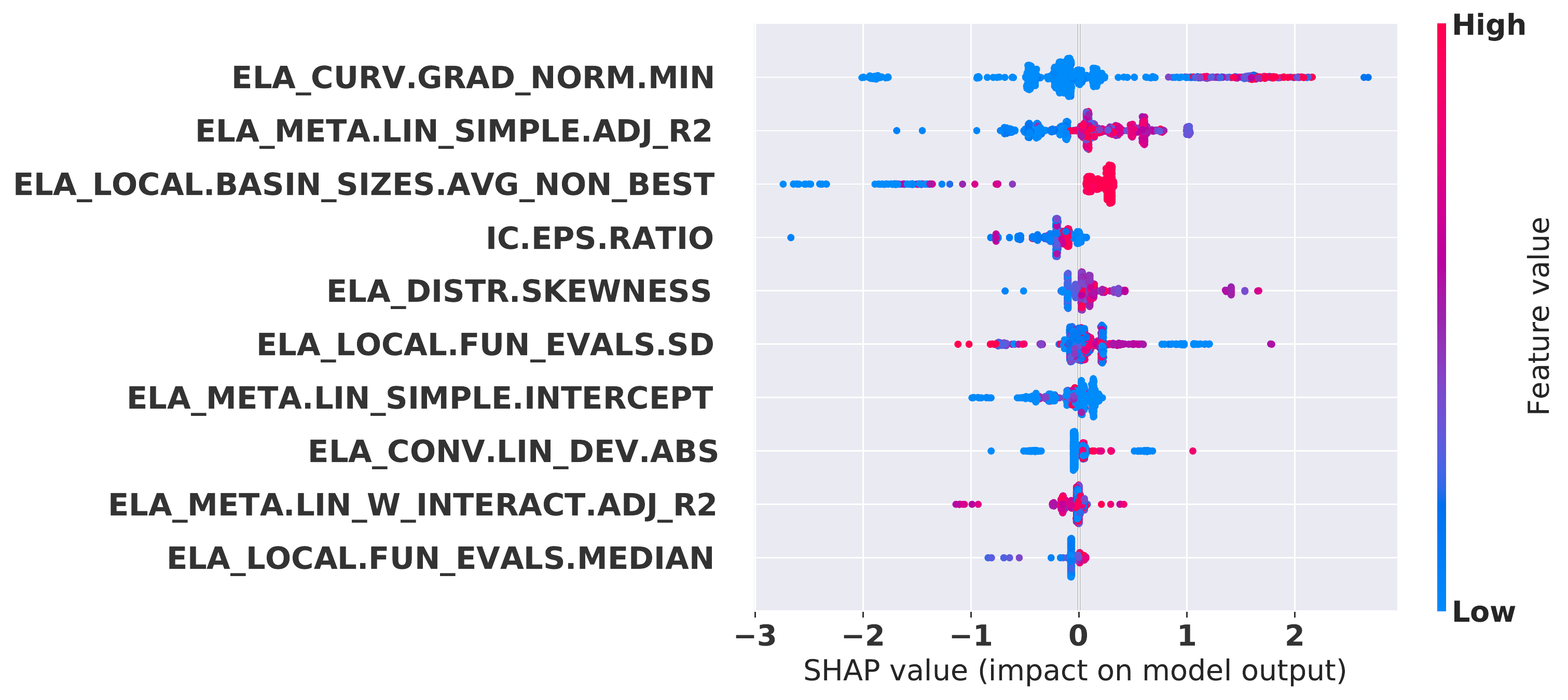} 
    \caption{Logarithmic transformation of the target precision STR DT.} 
    \label{fig7:b} 
    \vspace{4ex}
  \end{subfigure} 
  \begin{subfigure}[b]{0.5\linewidth}
    \centering
    \includegraphics[width=0.75\linewidth]{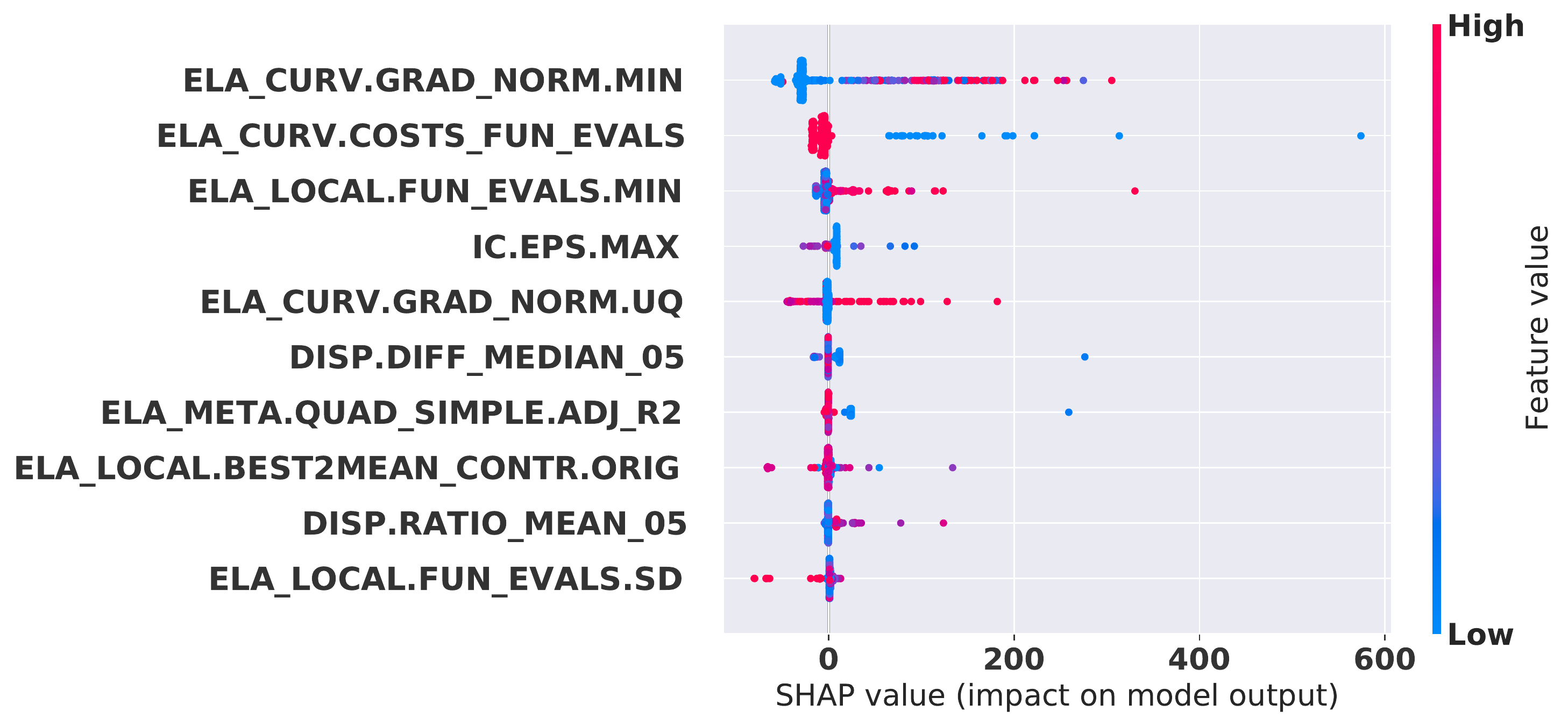} 
    \caption{Target precision MTR DT.} 
    \label{fig7:c} 
  \end{subfigure}
  \begin{subfigure}[b]{0.5\linewidth}
    \centering
    \includegraphics[width=0.75\linewidth]{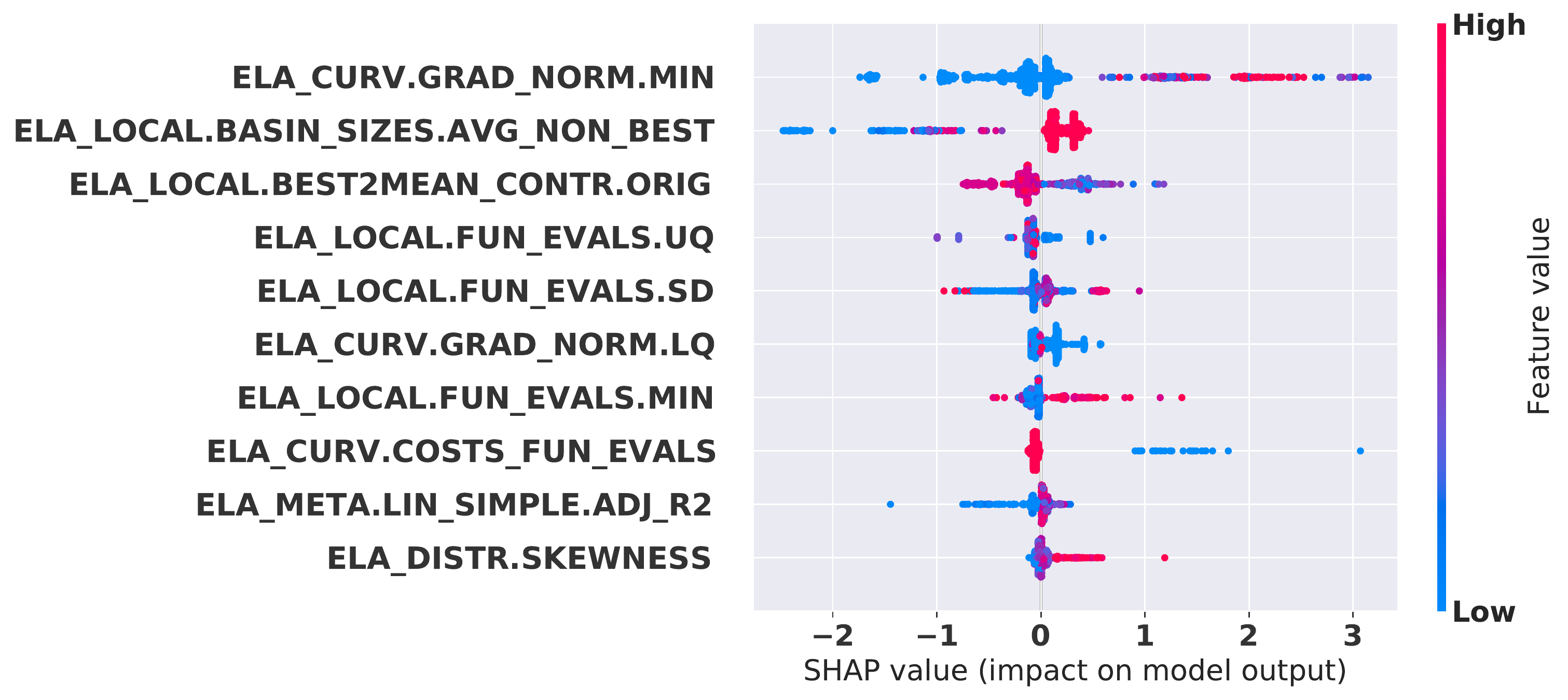} 
    \caption{Logarithmic transformation of the target precision MTR DT.} 
    \label{fig7:d} 
  \end{subfigure} 
  \caption{SHAP value impact of the STR and MTR DT models.}
  \label{fig:SHAP_DT} 
\end{figure*}

Figure~\ref{fig:SHAP_DT} presents the SHAP value impact of the STR and MTR DT models, both for the original target precision and its logarithmic transformation. The plots presented in this figure present the positive and negative relationships with the original target precision or its logarithmic transformation. The dots presented in the plots correspond to all instances from the training data set (i.e., in our case the first fold). The descending order of the ELA features presents their importance starting from the most important one. The colors used are related to the magnitude of the ELA feature value, where higher values are red and lower value are blue. The impact of the ELA feature value to the target variable prediction is its horizontal location.

Using the figure and focusing on the prediction of the original target precision reached by the DT models in the STR and MTR scenario, the models differ in one from 10 most important ELA features. The STR model uses the ``$ela\_local.fun\_evals$", while the MTR model uses ``$ela\_distr.number\_of\_peaks$". The patterns of the other nine shared features are almost the same in the STR and the MTR scenario. For example, if we have a high value of the ``$ela\_curv.gra\-d\_norm.min$", it contributes by adding a large value to the original target precision reached. The feature ``$ela\_curv.gra\-d\_norm.min$", which is most important feature in most of the scenarios, represents the aggregation of minimum values of the gradients length in all the runs when searching for the optimal solution.This means that high value ``$ela\_curv.gra\-d\_norm.min$" is an indication of difficulty in solving the benchmark problem because it takes us away from the optimum reached with adding a large value on the target precision (i.e., error). In addition, looking at the ``$ela\_curv.grad\_norm.uq$", which represents aggregation of the upper quartile of the gradients lengths, lower values do not take us away from the target precision reached, while higher values of this ELA feature can decrease or increase the reached target precision, showing that the benchmark problem is difficult to be solved there. In the future, such kind of analysis can allow us to estimate and rank the problem difficulty concerning the values of the ELA features and their Shapley values. Looking at the most important features it seems that they all come from the classical ELA features group, except one that comes from the information content group.
Focusing on the logarithmic transformation of the original target precision reached used as a target variable in the STR and MTR scenario by DT models, it seems that most of the features are overlapping. These models also utilize one feature from the nearest-better clustering group. The impact ELA patterns of both models are very similar for each ELA feature separately. 

Comparing the STR DT models for both performance targets (i.e., the original target precision and its logarithmic transformation), it is obvious that the ELA features importance is different and there is overlapping in a small number of features.

In the case when RF is used to learn the predictive model, the STR and MTR scenario differ in the ELA features they utilize, for both the original target precision reached and its logarithmic transformation. They differ in five out of the 10 most important ELA features. Similar to the DT models, the most utilized features are the classical ELA features. The ``$ela\_curv.grad\_norm.min$" is the most important ELA feature for both DT and RF models in STR and MTR learning scenario. Comparing the DNN STR and MTR models concerning the target precision, the importance of the ELA features is similar. They are overlapping in nine out of the 10 most important features. In case of the logarithmic transformation prediction, they are overlapping in five out of the 10 most important features. Compared to the RF and DT models, the DNN models utilized also features from the information content and nearest neighbours groups. Figures about the RF and DNN scenarios are available at out GitHub repository.

Figure \ref{fig:SHAP_DT} presents the explanations obtained from the training data from the first fold. To see if these results are consistent across different folds (i.e., if the features importance is consistent within a model across the folds) in case when DT, RF, and DNN are used, we have represented each ML model in each learning scenario as a vector of 99 Shapley values. For this purpose, we averaged the Shapley value for each ELA feature across all problem instances that belong to each training data fold. The vector with the averaged Shapley values is the ML model representation. 

\begin{figure*}[tb]
    \centering
    \includegraphics[scale=0.3]{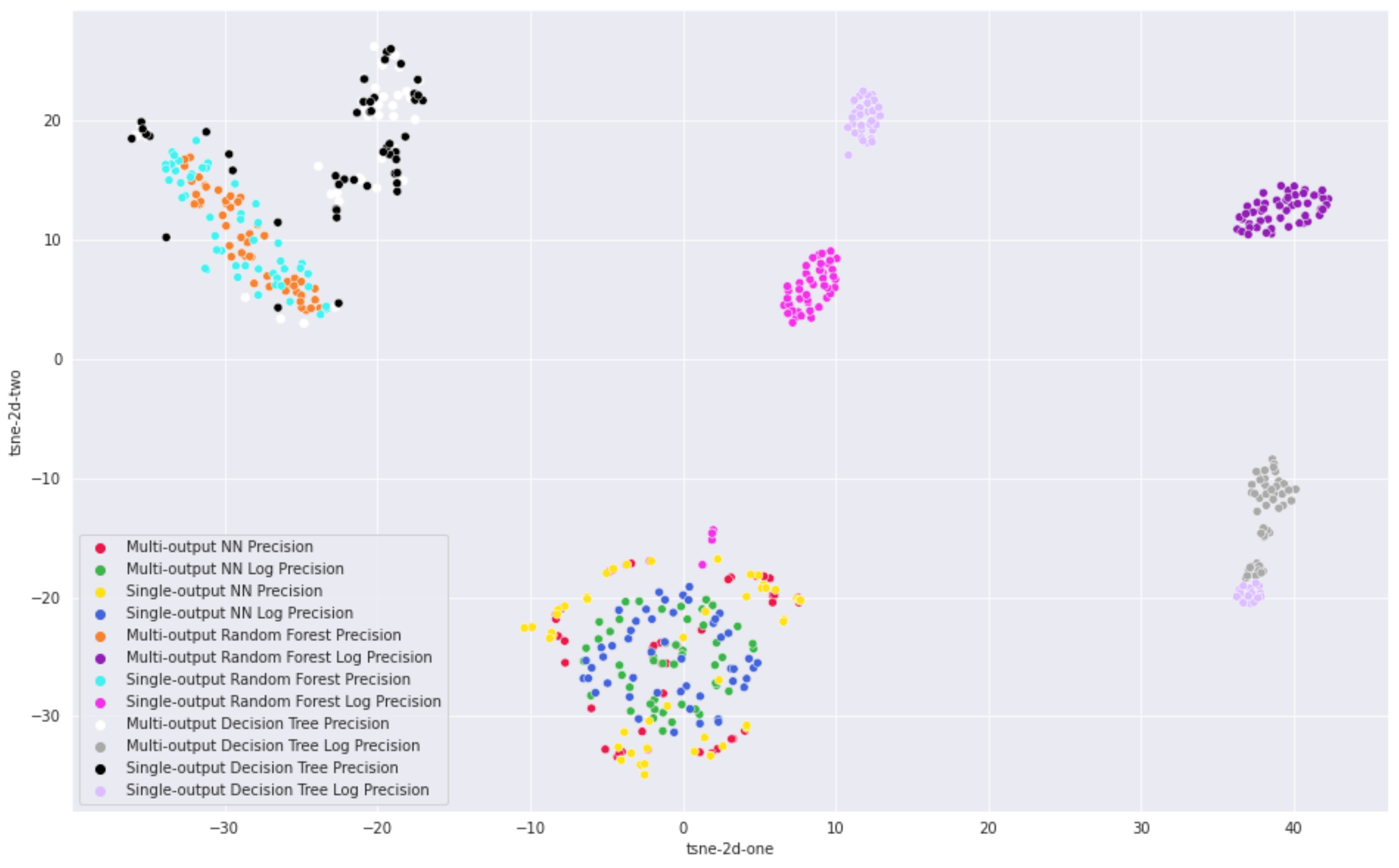}
    \caption{t-SNE visualization of the STR and MTR models for original target precision and its logarithmic transformation trained per each fold. The models are represented as vectors of 99 Shapley values (i.e., one per each ELA feature).}
   \label{fig:tsne}
\end{figure*}

\begin{figure*}[tb]
    \centering
    \includegraphics[scale=0.3]{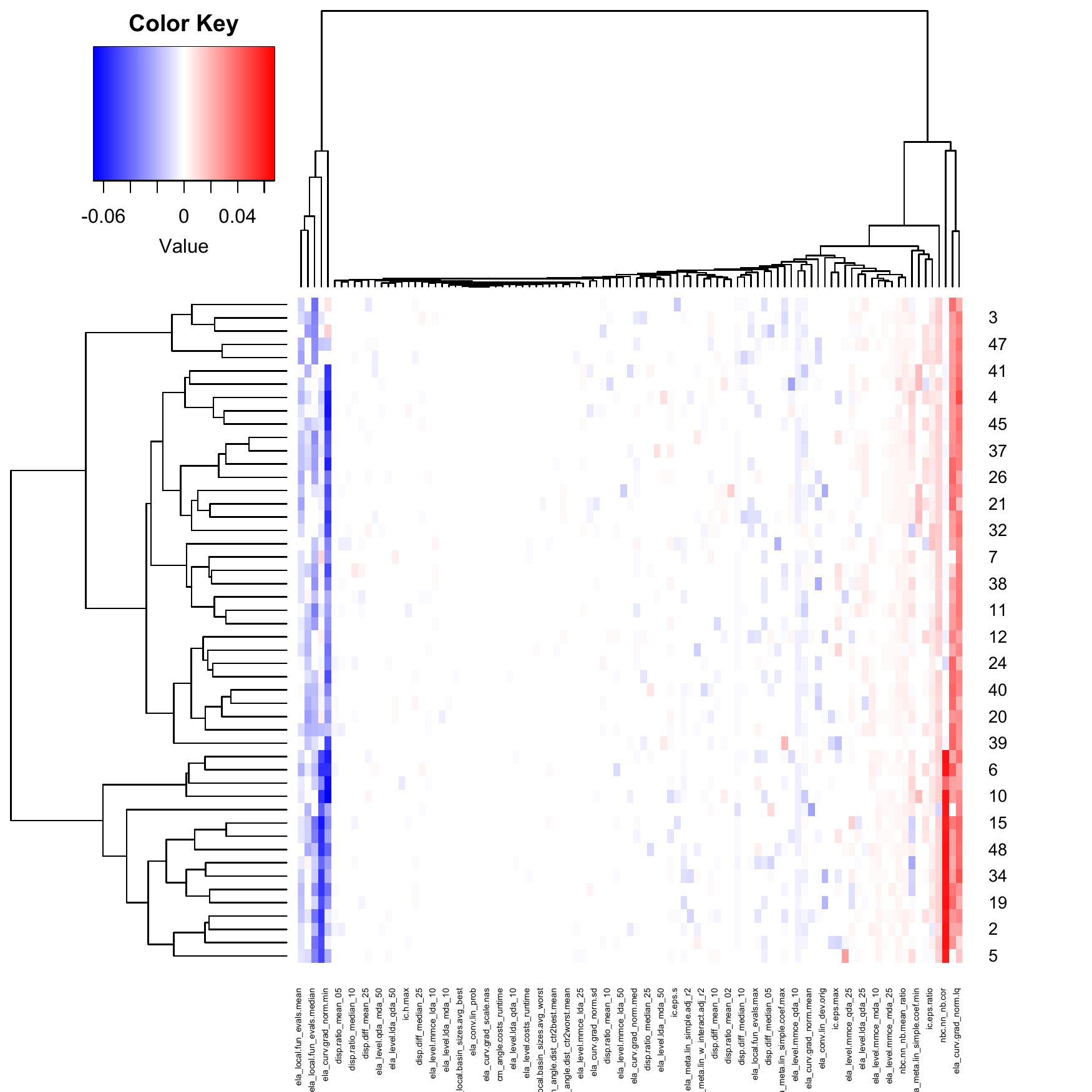}
    \caption{A heatmap where the DT models trained on the 50 folds for predicting the logarithmic transformation are reordered concerning how they utilized the ELA features.}
   \label{fig:clust}
\end{figure*}

Figure~\ref{fig:tsne} presents t-SNE visualization of each ML model in two-dimensional space using their Shapley representation. We have used the default parameters from the t-SNE visualization available from the python package \emph{scikit-learn}~\cite{pedregosa2011scikit}. Looking at the figure, we can assume that the models trained by the same ML algorithm across different folds are consistent since their Shapley representations place them close together. This result indicates that no matter which fold is used, the ELA features importance utilized by the models is similar. The only exceptions are the STR DT models trained for predicting the logarithmic transformation of the reached target. These results indicate that the same DT model trained on different folds utilized different ELA features (i.e., their importance changed). To check this, Figure~\ref{fig:clust} presents a heatmap with dendrogram, where the DT models trained on the 50 folds for predicting the logarithmic transformation are reordered concerning their Shapley values representation (i.e., 99 Shapley values, one per each feature). Looking at the dendrogram presented in the rows, it is obvious that these models are split into two clusters according to the information of the ELA feature importance, which is actually the result presented using the t-SNE visualization. This happens because different folds consist of different instances from the same problems that are obtained with different random transformations. These results indicate that the DT model is less robust to variations that exist between the ELA values across instances from the same problem when predicting the logarithmic transformation. 

\begin{figure*}[tb] 
  \begin{subfigure}[b]{0.5\linewidth}
    \centering
    \includegraphics[width=0.5\linewidth]{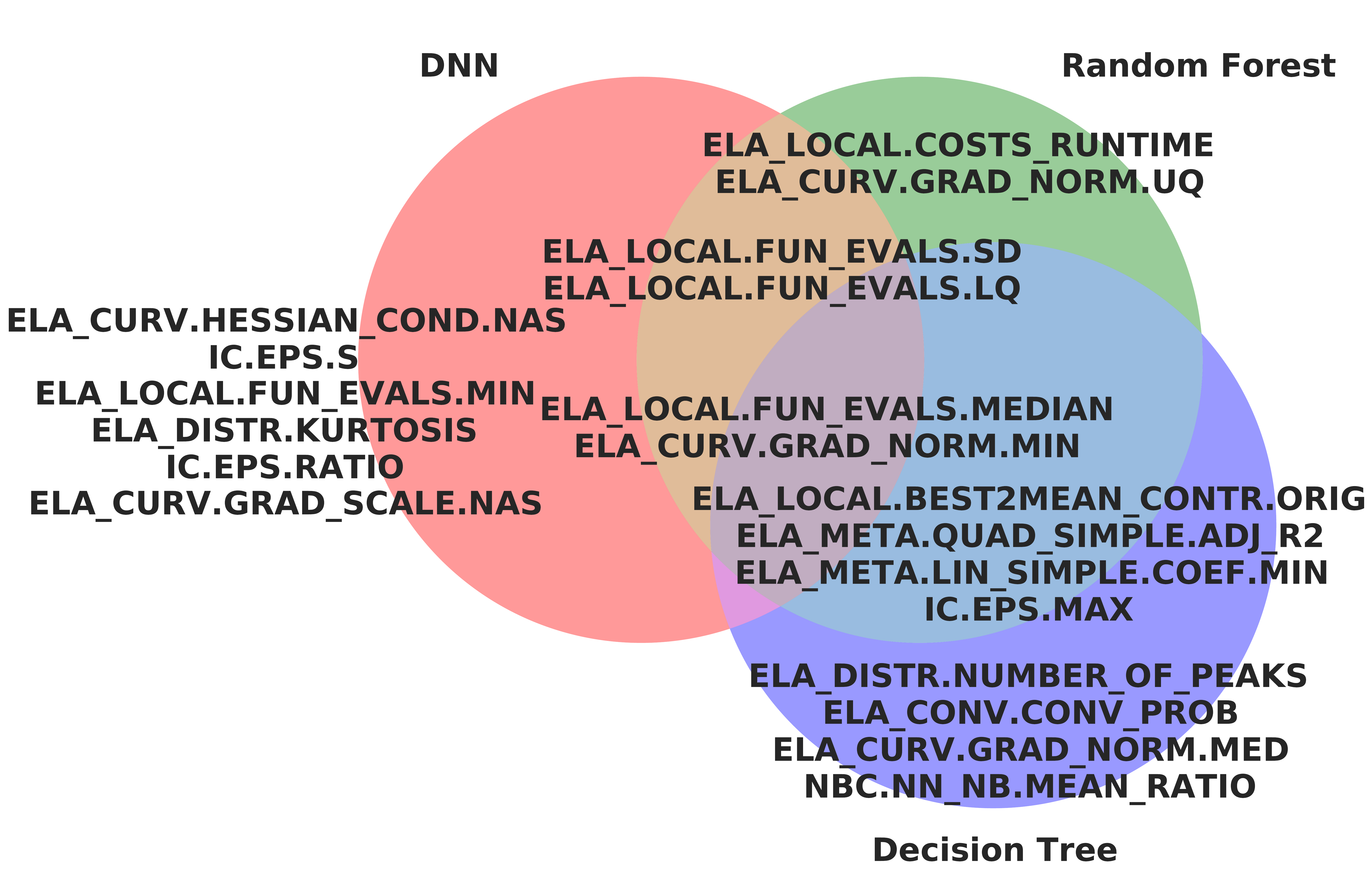} 
      \caption{STR target precision.} 
    \label{fig10:a} 
    \vspace{4ex}
  \end{subfigure}
  \begin{subfigure}[b]{0.5\linewidth}
    \centering
    \includegraphics[width=0.5\linewidth]{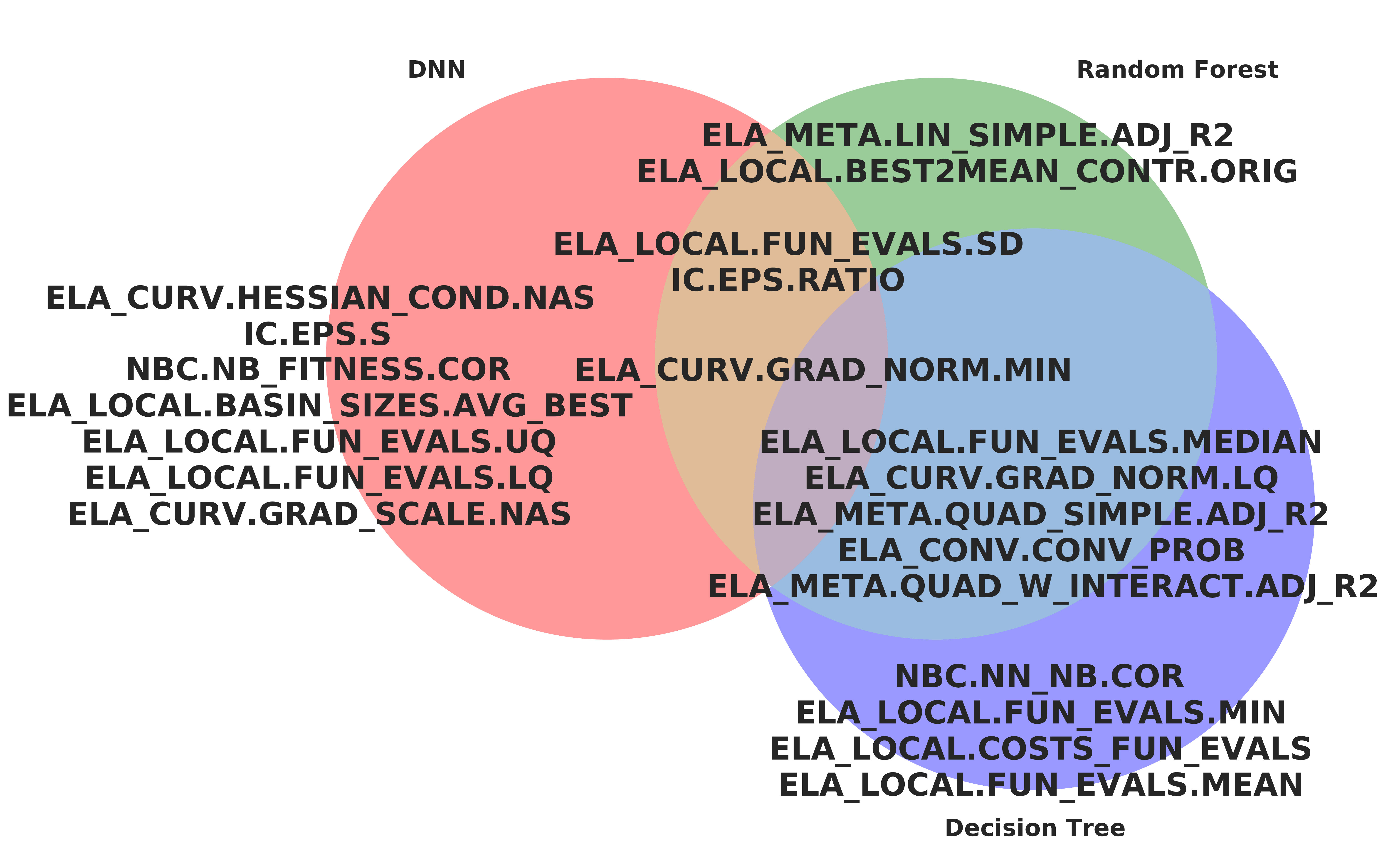} 
    \caption{STR logarithmic transformation.} 
    \label{fig10:b} 
    \vspace{4ex}
  \end{subfigure} 
  \begin{subfigure}[b]{0.5\linewidth}
    \centering
    \includegraphics[width=0.5\linewidth]{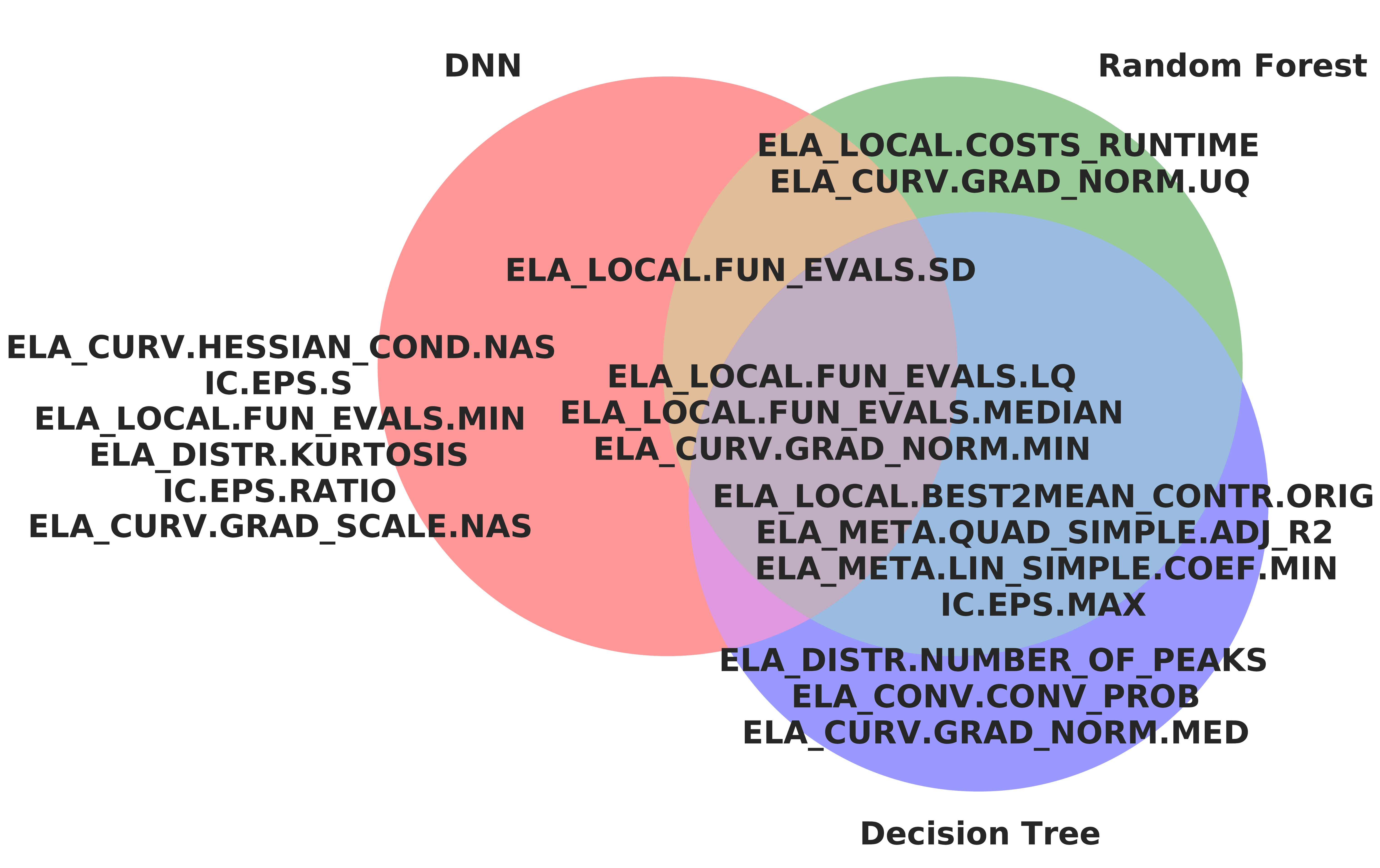} 
    \caption{MTR target precision.} 
    \label{fig10:c} 
  \end{subfigure}
  \begin{subfigure}[b]{0.5\linewidth}
    \centering
    \includegraphics[width=0.5\linewidth]{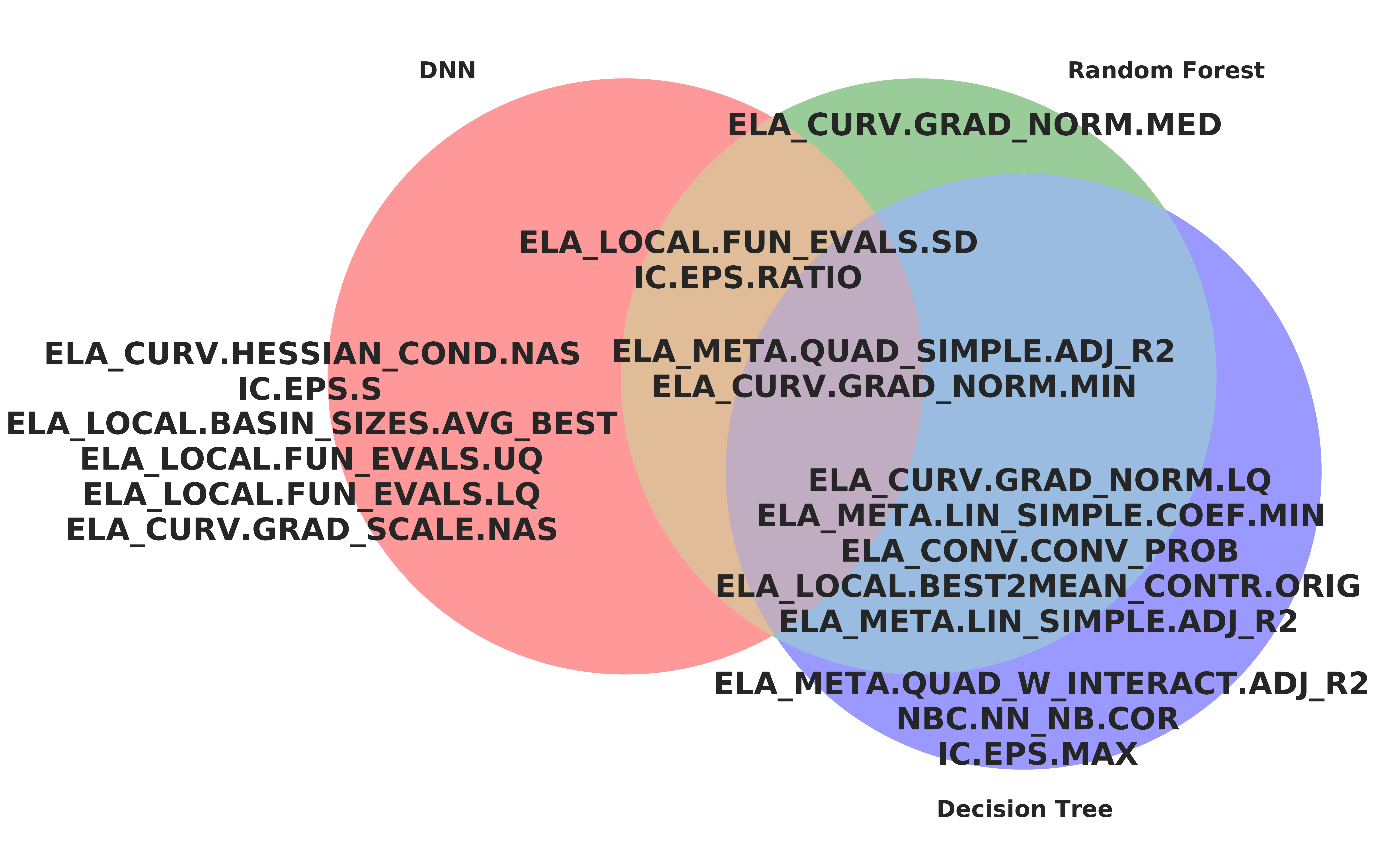} 
    \caption{MTR logarithmic transformation.} 
    \label{fig10:d} 
  \end{subfigure} 
  \caption{Intersection between the top 10 most important ELA features utilized by the DT, RF, and DNN models in STR and MTR learning scenario.}
  \label{fig:SHAP_Venn} 
\end{figure*}

To compare which ELA features are utilized across different ML models (i.e., DT, RF, and DNN), Figure~\ref{fig:SHAP_Venn} presents the intersections between the top 10 most important ELA features utilized by the DT, RF, and DNN models in STR and MTR learning scenario separately. For this purpose, the vectors for each ML model in each scenario used in the t-SNE visualization have been averaged within the scenario (e.g., STR and target precision, etc.). Based on the averaged Shapley values, the top 10 ELA features for each model and each scenario have been selected and used for the intersection analysis. From the figure, it follows that the tree-based models (DT and RF) share more similar ELA features. When we are comparing them to DNN models, the DNN models are most similar to the RF models and only few ELA features are overlapping with the DT models. The $``ela\_curv.grad\_norm.min"$ is the ELA feature that belongs to the 10 most important features for every ML algorithm from our portfolio no matter whether it is the STR or MTR scenario. In the future, the union of these features could be reused as a general feature selection to train a ML model.

\begin{figure*}[!ht] 
  \begin{subfigure}[b]{\textwidth}
    \centering
    \includegraphics[width=0.5\textwidth]{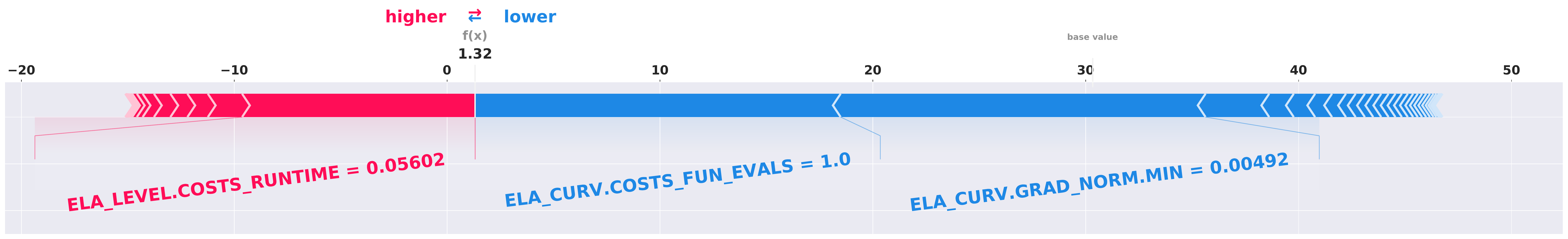} 
      \caption{DT prediction.} 
    \label{fig13:a} 
    \vspace{4ex}
  \end{subfigure}
  \begin{subfigure}[b]{\textwidth}
    \centering
    \includegraphics[width=0.5\textwidth]{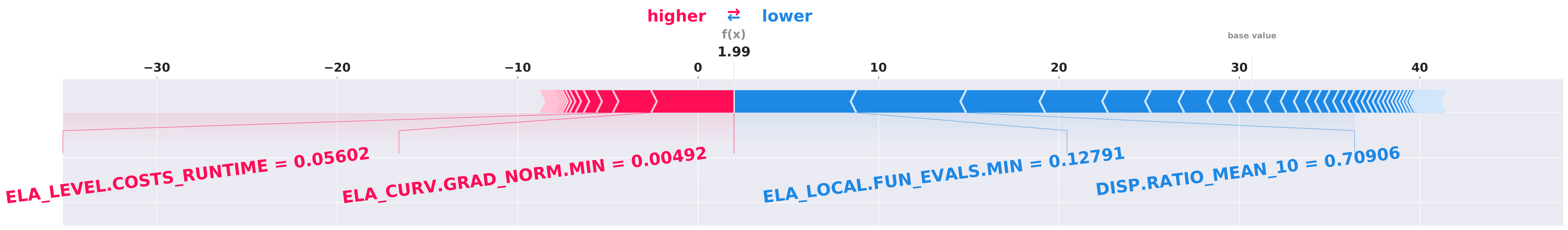} 
   \caption{RF prediction.} 
    \label{fig13:b} 
 \vspace{4ex}
  \end{subfigure} 
  \begin{subfigure}[b]{\textwidth}
    \centering
    \includegraphics[width=0.5\textwidth]{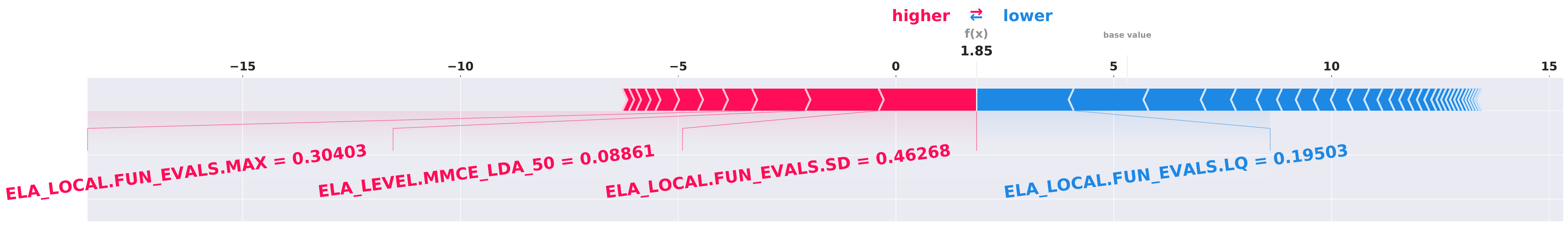} 
   \caption{DNN prediction.} 
    \label{fig13:c} 
 \vspace{4ex}
  \end{subfigure} 
  \caption{SHAP value impact of the STR models to explain the predicted target precision reached on the first instance of the 11th benchmark problem.}
  \label{fig:SHAP_DT_single} 
\end{figure*}
\subsubsection{Local explanations}
After presenting the global impact of the ELA features concerning different supervised ML regression models, here the focus is on local explainability. To provide local explanations, the first instance of the 11th benchmark problem from the first fold was selected. This was done with the purpose to see what makes the prediction of this problem difficult using DT or RF (i.e., which ELA feature influence here) and why the DNN model has lower error compared to the other two models.

\begin{figure}
    \centering
    \includegraphics[scale=0.15]{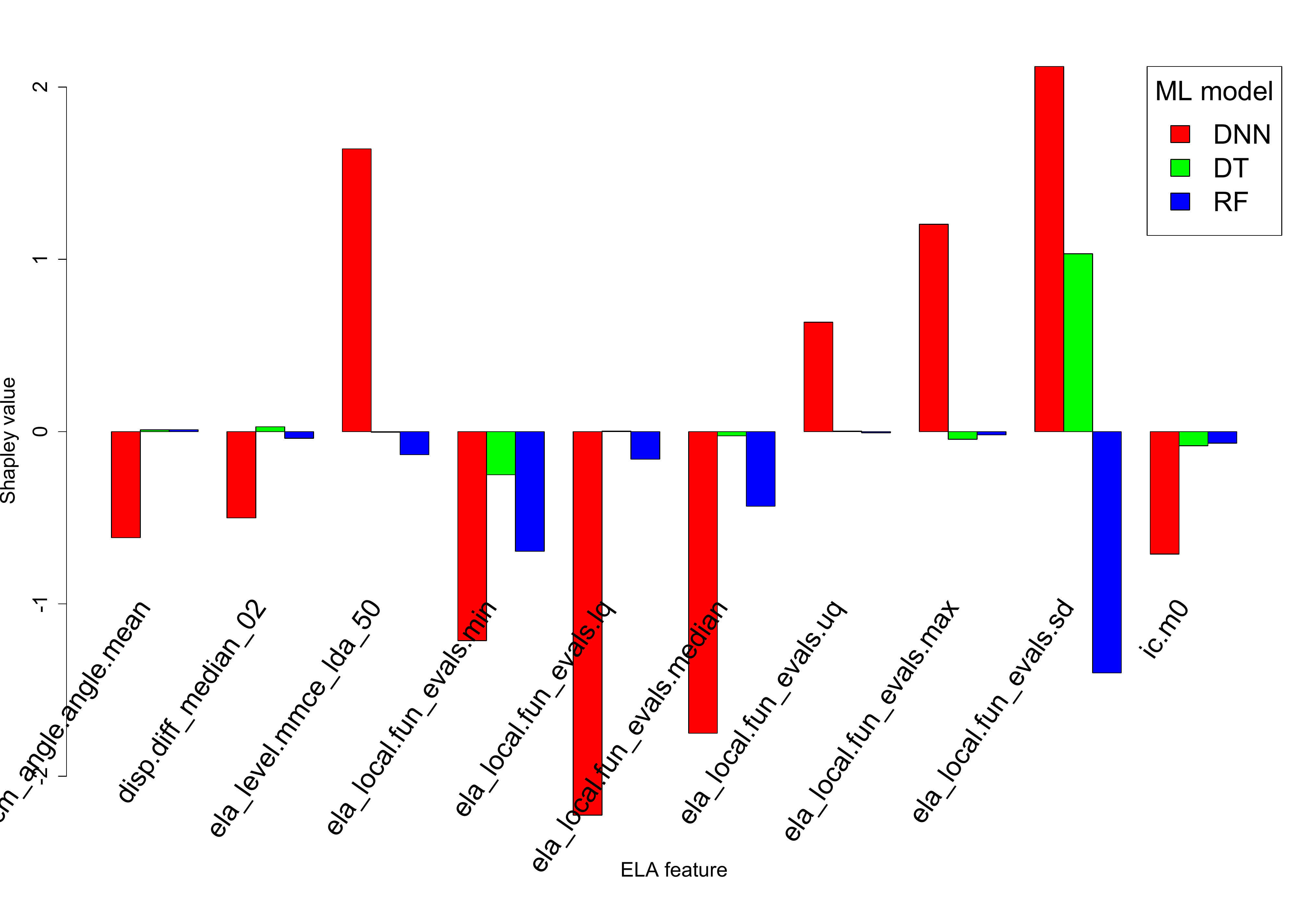}
    \caption{The top 10 most important ELA feature utilized by the DNN on the first instance of the 11th benchmark problem.}
    \label{fig:feature_importance}
\end{figure}

Figure~\ref{fig:SHAP_DT_single} presents the SHAP impact of the 10 most important ELA features that contribute to the original target prediction for the selected instance by the DT, RF, and DNN models in STR scenario. Form the figure, it is obvious that using the most important features in the cases of the DT and RF, the range of the prediction is from -15 to 45. However, the DNN regressor utilizes the features by making the prediction in the range from -5 to 13, so smaller prediction errors are achieved.  To go in more detail, Figure~\ref{fig:feature_importance} presents the impact of the top most important 10 ELA features utilized by the DNN model on this instance, together with their importance when the DT or RF models are used. From the figure, it follows that the importance of these 10 features is not the same when the DT and RF are used. These results indicate that the selection of the ELA features portfolio is also dependent on which supervised ML method will be used.

\section{Conclusion}
In this study, we investigated the ELA features importance in automated algorithm performance prediction when different supervised ML regression methods are utilized. Evaluating the learning task on the 50 instances from the 24 COCO benchmark problems and one CMA-ES configuration by using decision tree, random forest, and deep neural network, it follows that in this learning scenario the most important ELA features are coming from the classical \emph{ela} group. The calculated Shapley values were used to estimate the contribution of each ELA feature to the performance prediction. The experimental results showed that a different set of ELA features are important for different problem instances depending on which supervised ML method is utilized. This indicates that selection of ELA features is very dependent on the ML method that is utilized for learning the predictive model. So, depending on which supervised ML method is used, the impact of the ELA features to the model prediction performance changes. 

For future work, we are planning to analyse different families of optimization algorithms to investigate which ELA features are related to their exploration and exploitation capabilities, and further recommend the most appropriate ML method that can be used to perform automated algorithm performance prediction.

\vspace{1ex}
{\scriptsize{
\textbf{Acknowledgments.} 
This work was supported by projects from the Slovenian Research Agency: research core funding No. P2-0098 and projects No. Z2-1867 and N2-0239.
}
%
%

%
\bibliographystyle{splncs04}
\bibliography{main}

\end{document}